%% file: ms.tex
\documentclass[11pt]{article} 

\usepackage{graphicx}
\usepackage{booktabs}
\usepackage{algorithm}
\usepackage{algpseudocode}
\usepackage[T1]{fontenc}
\usepackage{lmodern}
\usepackage[flushleft]{threeparttable}
\usepackage[hidelinks]{hyperref} %
\hypersetup{hidelinks}
\usepackage{multirow}
\usepackage{longtable}
\usepackage[margin=3cm]{geometry} %
\usepackage{amsmath}
\usepackage{authblk}
\usepackage{caption}

\title{Exploring possible vector systems for faster training of neural
networks with preconfigured latent spaces}
\author[1]{Nikita Gabdullin}
\affil[1]{Joint Stock "Research and production company "Kryptonite" \authorcr
E-mail: n.gabdullin@kryptonite.ru}
\date{}
\begin{document}

    \captionsetup[table]{labelformat={default},labelsep=period,name={Table}}

    \maketitle

    \begin{abstract}
        The overall neural network (NN) performance
        is closely related to the properties of its embedding distribution in
        latent space (LS). It has recently been shown that predefined vector
        systems, specifically \emph{A\textsubscript{n}} root system vectors, can
        be used as targets for latent space configurations (LSC) to ensure the
        desired LS structure. One of the main LSC advantages is the possibility
        of training classifier NNs without classification layers, which
        facilitates training NNs on datasets with extremely large numbers of
        classes (\emph{n\textsubscript{classes}}). This paper provides a more
        general overview of possible vector systems for NN training along with
        their properties and methods for vector system construction. These
        systems are used to configure LS of encoders and visual transformers to
        significantly speed up ImageNet-1K and 50k-600k classes LSC training. It
        is also shown that using the minimum number of LS dimensions
        (\emph{n\textsubscript{min}}) for specific \emph{n\textsubscript{classes
        }}results in faster convergence. The latter has potential advantages for
        reducing the size of vector databases used to store NN embeddings.
  
    \end{abstract}

    
    \input{full}

\end{document}

%% file: full.tex
\section{Introduction}
\label{introduction}

Rapid spreading of neural networks (NNs) over the last decade has
increased the demand for NNs capable of producing high-accuracy
predictions for unprecedented amounts of unseen data. More and more
applications require multi-domain capabilities like, for instance,
simultaneously working with images and text, or sound and text, etc~\cite{clip,Multimodal}. 
This is achieved by projecting data of different
domains into the same NN latent space (LS). As in case of single-domain
data, the overall NN performance is closely related to the properties of
its embedding distribution. This has inspired researchers to propose
methods that take LS properties into consideration during training and
inference~\cite{CenterLoss, lsconf, SphereFace}.

It has previously been proposed that identifying key LS properties and
using vector systems with similar properties for LS configuration (LSC)
can allow one to train classifier NNs which have no classification
layers~\cite{LSC}. This allows using the same NN architecture for
datasets with large and even variable numbers of classes
(\emph{n\textsubscript{classes}}). The configuration used in that study
corresponded to root system \emph{A\textsubscript{n }}which has very
well-spaced vectors used as targets for cluster centers of NN embedding
distributions. However, \emph{A\textsubscript{n}} interpolation is
required to obtain a sufficiently large number of vectors
(\emph{n\textsubscript{vects}}) for reasonable LS dimension
(\emph{n\textsubscript{dim}}) on datasets with large
\emph{n\textsubscript{classes}}. In this paper we study methods for
constructing other vector systems with a desired set of properties which
do not require interpolation to accommodate a large number of vectors
while having an acceptable vector spacing. These vector systems are used
to train NNs to attain LSC features previously summarized in Section 6
in~\cite{LSC} (references to Sections in~\cite{LSC} are hereafter
referred to using forward slashes, e.g. Section /6/).

The rest of the paper is organized as follows: Section~\ref{Vn} provides a
framework for vector system search and their properties' estimation,
Section~\ref{Exp} experimentally verifies the feasibility of vector system
training and compares it with the conventional Cross-Entropy (CE) loss
training, Section~\ref{discussions} discusses the implications of the experimental
results, and Section~\ref{conclusions} concludes the paper.

\section{Vector systems for latent space configuration}
\label{Vn}

\subsection{Obtaining vector systems through base vector
coordinates permutations}
\label{baseV}

For the purposes of this work, we define vector systems
\emph{V\textsubscript{n}} as sets of unique \emph{n}-dimensional vectors
obtained using specific rules, or generating functions, for a family of
individual vectors \emph{v}

\begin{equation}
	V_{n} = f_{gen}(n) = set\left( v_{i} \right),\ i = 1...n_{vects},
	\label{eq:V_n}
\end{equation}
\unskip

where \emph{n\textsubscript{vects}} is the number of vectors in the
system. We are primarily interested in vector systems with a large
number of vectors which properties could facilitate fast NN training and
good inference performance. It has been previously shown that one of the
most important properties of vector systems is the separation between
vectors used as training targets for NN embedding cluster centers. 
Hence, we will use \emph{n\textsubscript{vects}} and minimum
cosine similarity (\emph{mcs}) as criteria for assessing the suitability
of the vector system. The latter is defined as

\begin{equation}
	mcs = \min(abs(cossim(v_{i},v_{j}))),\ i \neq j,\ v\  \in \ V_{n}.
	\label{eq:msc}
\end{equation}
\unskip

As mentioned above, the main target is finding \emph{V\textsubscript{n}}
with large \emph{n\textsubscript{vects }}and low \emph{mcs}. It has been
shown that NN training becomes complicated when \emph{mcs} approaches 0.9~\cite{LSC},
so we obtain the following inequality

\begin{equation}
	0.5 < mcs\  \ll 0.9,
	\label{eq:msc_range}
\end{equation}
\unskip

which uses \emph{A\textsubscript{n}} vector spacing as the lower bound.

In general, vector systems can be constructed by choosing some base
vector and obtaining the complete system as its unique permutations~\cite{Igor}. 
Table~\ref{tab:21} shows that, for instance, root system
\emph{A\textsubscript{n-1}} can be obtained as permutations of
\emph{n}-dimensional base vector with coordinates {[}1,0...0, -1{]}.
Similarly, one can obtain vectors corresponding to vertices of
permutohedron \emph{P\textsubscript{n-1}}. \emph{A\textsubscript{n}} and
\emph{P\textsubscript{n}} correspond to the least and the most
vector-rich systems in \emph{n} dimensions. Table~\ref{tab:21} also shows that
\emph{A\textsubscript{n}} has the lowest number of evenly spaced vectors
with \emph{mcs} independent of \emph{n}. On the contrary,
\emph{P\textsubscript{n}} allows obtaining an enormous number of vectors
which angular spacing rapidly decreases with \emph{n}. This limits its
application in NN training, as further discussed in Section~\ref{Exp}.

Constructing new vector systems can be performed by adding non-zero
elements to \emph{A\textsubscript{n}} base vector. In terminology of Lie
algebras and root systems, such vectors correspond to various weight
vectors whose weight orbits constitute complete vector systems~\cite{Lie}. 
Unfortunately, there is no convenient general notation for
such systems. In this paper we use label \emph{D} (due to its relation
to Dynkin index and Dynkin label~\cite{LieART}) which reflects the
numbers of unique non-zero elements, so, for instance,
\emph{A\textsubscript{n}} label is ``11'' (one 1 and one -1). The
general vector system notation then becomes \(V_{n}^{D}\). Whereas this
obviously can lead to confusing notations for vector systems with many
non-zero elements, we will use these labels for simple vector systems. 
It should be stressed that all \(V_{n}^{D}\) vectors are
\emph{n}-dimensional, which differs from the standard root system
notation where \emph{A\textsubscript{n}} root system has
(\emph{n+1})-dimensional vectors, and practical application of
\emph{A\textsubscript{n}} vectors for NN training requires projecting
the original vectors back to \emph{n}-dimensional space~\cite{LSC}.

Using \emph{D}, one could also obtain \emph{n\textsubscript{vects}} for
arbitrary \(V_{n}^{D}\) by calculating unique permutations as

\begin{equation}
	n_{vects} = \frac{n!}{\lambda_{1}!\lambda_{2}!...\lambda_{k}!\left( n - \sum_{1}^{k}\lambda_{i} \right)!},
	\label{eq:n_vects}
\end{equation}
\unskip

where \(\lambda_{1}...\lambda_{k}\) are elements of \emph{D}. This
equation can also be used to find minimum \emph{n} required to
accommodate the desired \emph{n\textsubscript{vects,}} which we will
refer to as \emph{n\textsubscript{min}.} It can be found iteratively
from (\ref{eq:n_vects}) since solving this equation for arbitrary \emph{D} is not
practical.

In particular, we will be interested in two vector systems
\(V_{n}^{21}\) and \(V_{n}^{22}\) which correspond to
\emph{A\textsubscript{n}} base vector with extra 1, and {[}1,-1{]},
respectively\footnote{Note that numbers in \emph{D} should be treated as
separate digits, so \(V_{n}^{21}\) reads as \emph{V\textsubscript{n}}
``two-one'', not ``twenty one''. Whereas this distinction is very
important, it holds true only for numbers below 10. Therefore, this
notation cannot be applied universally.}. These simple operations
allow a remarkable increase in \emph{n\textsubscript{vects}} accompanied
by an acceptable increase in \emph{mcs}. As next Sections will show,
these vector systems will prove to be extremely useful for NN training.

\begin{table}
	\caption{Properties of LS configurations used in this study.} 
	\label{tab:21}
	\centering
	\begin{tabular}{|c|c|c|c|c|}
	  \hline
	  Vector system & Base vector & \emph{mcs} & {n\textsubscript{vects}} (n=384) & {n\textsubscript{vects}} eq. (\ref{eq:n_vects}) \\ \hline
		A\textsubscript{n-1} (\(V_{n}^{11}\)) & {[}1,0...0,-1{]} & 0.5 & 147k
		& a = n(n-1) \\ \hline
		\(V_{n}^{21}\) & {[}1,1,0...0,-1{]} & 0.67 & 28m & b =
		a(n-2)/2 \\ \hline
		\(V_{n}^{22}\) & {[}1,1,0...0,-1,-1{]} & 0.75 & 5.3b &
		b(n-3)/2 \\ \hline
		\emph{P\textsubscript{n-1 }}(\(V_{n}^{1...1}\)) & {[}n-1,n-2...1,0{]}* &
		1 - n\textsuperscript{-1} (approx.) & 10\textsuperscript{827} & n! \\ \hline
	\end{tabular}
	\\ \small *for consistency with other expressions, this conventional base vector
	form  \\ should  be  shifted by -\emph{n/2} so the resulting vector system is
	centered around zero.
\end{table}

\subsection{Empirical rules for constructing V\textsubscript{n} base vectors}
\label{Vn_rules}

Whereas Table~\ref{tab:21} shows that \(V_{n}^{21}\) and \(V_{n}^{22}\) are
already promising for NN training, the general properties of vector
system construction have also been studied allowing to observe some
empirical trends.

Firstly, \emph{V\textsubscript{n}} geometric properties are affected by
the number of its non-zero elements, \emph{i.e.} its label \emph{D}.
Hence, \emph{V\textsubscript{n}} properties remain the same for varying
\emph{n} as long as \emph{D} does not change. This is the reason why
\emph{mcs} of \emph{A\textsubscript{n}}, \(V_{n}^{21}\), and
\(V_{n}^{22}\) are constant while \emph{P\textsubscript{n} mcs} depends
on \emph{n}.

Secondly, \emph{mcs} grows with absolute values of elements in the
vector. This directly follows from the properties of dot product
operation. Hence, minimum absolute values are preferred for maximum
vector separation. Lastly, the number of zeroes in base vectors should
be no less than the number of any unique non-zero element to maximize
\emph{n\textsubscript{vects}}. Because of that \(V_{n}^{21}\) is most
effective for \emph{n}\(\geq\)5, and \(V_{n}^{22}\) for
\emph{n}\(\geq\)6.

Summarizing the above, one obtains the following rules:

\begin{itemize}
\item
  Increasing the number of non-zero elements in fixed \emph{n} changes
  \emph{V\textsubscript{n}} geometry, increases
  \emph{n\textsubscript{vects}} and \emph{mcs}.
\item
  Having higher absolute value numbers increases \emph{mcs}.
\item
  There should be at least as many zeroes as the number of any other
  unique non-zero elements (the largest number in \emph{D}).
\end{itemize}

\section{Experiments}
\label{Exp}

In this paper the experimental procedure exactly follows the LSC
methodology described in Sections /3.1/ and /3.3/. Vector systems
described in Section~\ref{Vn} are used as targets for embedding cluster centers
determining the LS configuration. For that reason, the terms vector
system and LS configuration are used interchangeably hereafter. For all
experiments LS dimension of NN \emph{n\textsubscript{dim}} matches
\emph{n} of the chosen vector system \(V_{n}^{D}\). As mentioned above,
for \emph{A\textsubscript{n}} and \emph{P\textsubscript{n}} this
requires projecting (\emph{n+1})-dimensional vectors to
\emph{n\textsubscript{dim}} as described in Section /3.2.1/. That
Section also describes the reasoning and methods for
\emph{A\textsubscript{n}} vector interpolation, and how
\emph{A\textsubscript{nr}} is obtained from \emph{A\textsubscript{n}} by
randomly shuffling the vectors. For the description of NN model
architectures and datasets the reader is referred to Section /4.1/. In
some ViT-S experiments \emph{n\textsubscript{min}} LS dimension is
achieved by projecting the original 384-dimensional embedding to
\emph{n\textsubscript{min}} using one fully-connected layer.

\begin{figure}[b]
	\centering
	\includegraphics[scale=0.45]{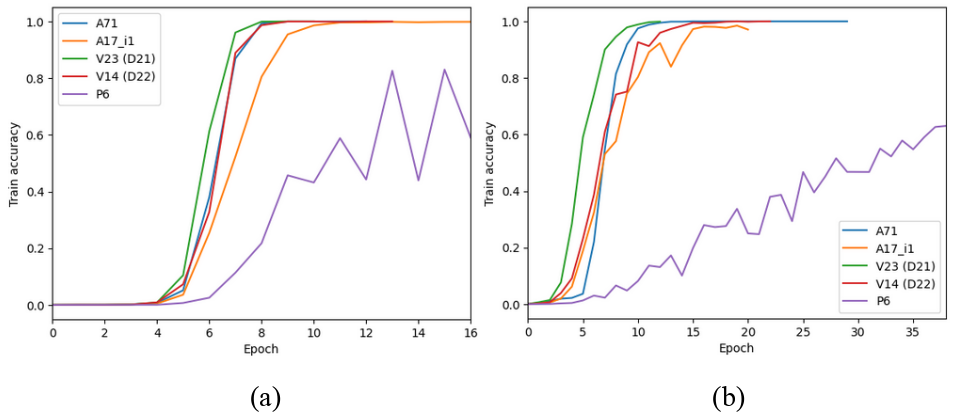} 
	\caption{Training speed of encoder model trained on \emph{n\textsubscript{classes}}=5000 dataset 
	with different LS configurations using (a) cosine and (b) Euclidean distances as loss functions.}
	\label{fig:311}
\end{figure}
\unskip

\subsection{Encoder training on 5k classes with different LS configurations}
\label{smalldim}

This Section studies the training speed of encoder model on a dataset
with \emph{n\textsubscript{classes}=}5000 obtained from ImageNet-1K
(i1k) depending on the LS configuration choice. Dimensions of all
configurations are chosen so \emph{n = n\textsubscript{min }}for 5k
classes. In all figures \emph{A\textsubscript{n\_i1}} refers to
\emph{A\textsubscript{n}} with one level of interpolation.
\emph{V\textsubscript{n}} (\emph{D21}) and \emph{V\textsubscript{n}}
(\emph{D22}) correspond to \(V_{n}^{21}\) and \(V_{n}^{22}\),
respectively.

Figure~\ref{fig:311} shows that \(V_{n}^{21}\) training is faster than training
using other configurations including \emph{A\textsubscript{n}}.
\emph{P\textsubscript{n}} training is the slowest and even diverges when
training with cosine loss, which is the main training loss for realistic
high-dimensional LS NNs~\cite{LSC}. This can be expected since
\emph{P\textsubscript{n} mcs} decreases with \emph{n}, as shown in Table~\ref{tab:21}. 
For that reason in the following Sections we focus on vector
systems which \emph{mcs} is independent of \emph{n}.

\subsection{NN training on large n\textsubscript{classes} datasets using different LS configurations}
\label{large_nclasses}

\begin{figure}[t]
	\centering
	\includegraphics[scale=0.40]{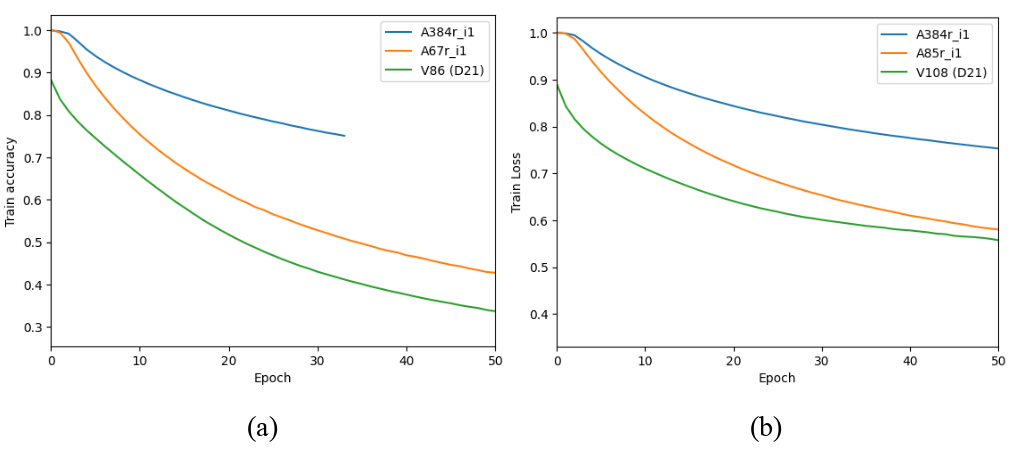} 
	\caption{Encoder training loss curves for various LS configurations
	for (a) \emph{n\textsubscript{classes}}=300k and (b) \emph{n\textsubscript{classes}}=600k training datasets.}
	\label{fig:321}
\end{figure}

\begin{figure}[t]
	\centering
	\includegraphics[scale=0.40]{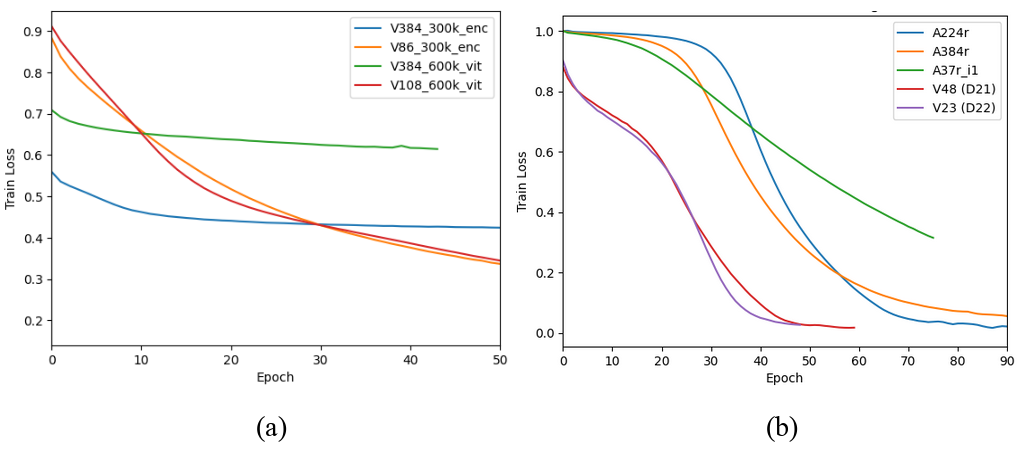} 
	\caption{The comparison of NN training for (a) encoder and ViT-S for \(V_{n}^{21}\) on different datasets 
	depending on \emph{n\textsubscript{dim}} (b) ViT-S with \emph{n\textsubscript{min}} and different 
	configurations for \emph{n\textsubscript{classes}}=50k.}
	\label{fig:322}
\end{figure}

Figure~\ref{fig:321} compares encoder training speed with \(V_{n}^{21}\) and two
\emph{A\textsubscript{nr}} configurations on 300k and 600k classes
datasets. In both cases \(V_{n}^{21}\) training is faster than
\emph{A\textsubscript{nr}} even when \emph{n\textsubscript{min}} is used
for \emph{A\textsubscript{nr}}. This shows that
n\emph{\textsubscript{min}} is desirable for LSC, since training with
default NN dimensions (i.e. 384 for ViT-S) is significantly slower.
Figure~\ref{fig:322} shows that this holds true for ViTs, too. It should also be
mentioned that \(V_{n}^{21}\) allows training with ten times higher
learning rate compared to interpolated \emph{A\textsubscript{n}} since
cluster centers are further apart.

Figure~\ref{fig:322} (b) shows that \emph{A\textsubscript{n}} interpolation
significantly slows down training compared to basic
\emph{A\textsubscript{n}} configuration, illustrating that
non-interpolated configurations are preferable. It also shows that
\(V_{n}^{22}\) training converges slightly faster than
\(V_{n}^{21}\) training. However, this difference is much smaller than
the one obtained by transitioning from \emph{A\textsubscript{n}} to
other configurations. Nevertheless, \(V_{n}^{22}\) can be very promising
for large \emph{n\textsubscript{classes}} training since it allows to
accommodate more vectors, as shown in Table~\ref{tab:21}.

\unskip

\unskip

\subsection{Comparing i1k training with LSC and conventional classifiers}
\label{i1k}

\begin{figure}[t]
	\centering
	\includegraphics[scale=0.45]{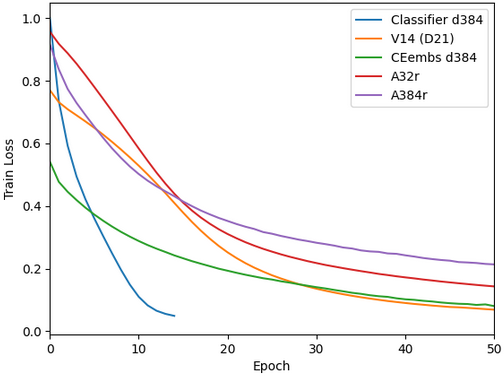} 
	\caption{Loss curves of ViT-S classifier trained with CE loss and
	ViT-S without classification layer trained with LSC using different LS configurations.}
	\label{fig:331}
\end{figure}

Figure~\ref{fig:331} shows ViT-S training results for LSC with
\emph{A\textsubscript{nr}}, \(V_{n}^{21}\), and \emph{CEembs}, and
conventional classification. The latter is obtained by training ViT-S
with classification layer for i1k \emph{n\textsubscript{classes}}=1000
using CE loss (CE loss values are normalized in Figure~\ref{fig:331} by the
maximum loss for clear comparison with other losses). For the complete
discussion about \emph{CEembs} the reader is referred to Section /4/. It
shows that comparing to the best LSC result from~\cite{LSC}
(\emph{A\textsubscript{384r}} training), a significant increase in
training speed is achieved by utilizing \(V_{14}^{21}\) as the target
configuration. Furthermore, \(V_{14}^{21}\) training becomes slightly
faster than \emph{CEembs} training for later epochs. It was previously
hypothesized that \emph{CEembs} correspond to the best possible
configuration, which in light of current results is clearly not the
case. Hence, it is possible to find better LS configurations than the
ones obtained from pretrained NNs.

It should be noted that LSC training is still slower than CE training
requiring about 3-4 times more epochs to achieve the same accuracy:
CE-trained classifier achieves 95\% accuracy on 13\textsuperscript{th}
epoch and LSC with \(V_{14}^{21}\) achieves similar values around
35\textsuperscript{th} epoch. However, LSC training can still be
improved by finding better configurations and training methods. This
paper studied vector systems from the perspective of \emph{msc} and
\emph{n\textsubscript{vects}}, though there might be other properties
which correspond to faster training. This can be illustrated by the fact
that whereas \(V_{n}^{21}\) \emph{mcs} is larger than that of
\emph{A\textsubscript{n}}, \(V_{n}^{21}\) configuration still allows
faster training, which indicates that \emph{mcs} relation to training
speed is non-trivial. There also might exist optimization techniques
more suitable for embedding matching training with cosine loss. Both of
these topics will be studied in greater detail in the future.

\unskip

\section{Discussions}
\label{discussions}

\subsection{Application of \emph{n\textsubscript{min}} approach to conventional classifiers}
\label{usual_nmin}

\begin{figure}[t]
	\centering
	\includegraphics[scale=0.55]{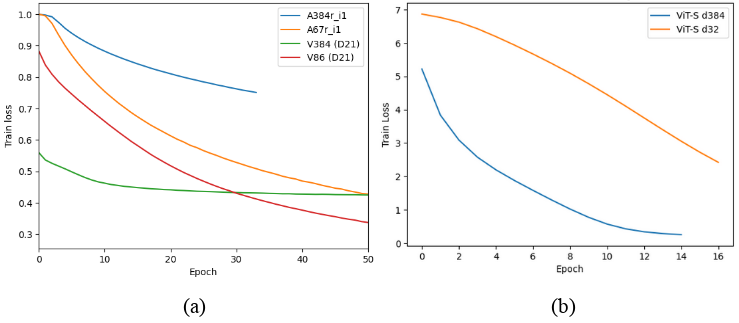} 
	\caption{Changes in ViT-S training speed when using
	\emph{n\textsubscript{min}} with (a) LSC and (b) a conventional classifier trained with CE loss.}
	\label{fig:411}
\end{figure}

Figure~\ref{fig:411} shows that whereas using \emph{n\textsubscript{min}} is
beneficial for LSC training, this effect does not occur for standard
classifiers trained with CE loss. Whereas \emph{n\textsubscript{dim}}=32
in Figure~\ref{fig:411} (b) is sufficient to allocate
\emph{n\textsubscript{classes}}=1000 well-separated i1k class clusters
in LS of that dimension, a higher-dimension classification layer can
achieve higher accuracy faster due to the ability to find clusters in a
less structured LS. Therefore, the \emph{n\textsubscript{min}} argument
is not directly applicable to conventional methods.

However, working with NN classifiers which have no classification layers
is the key property which facilitates large
\emph{n\textsubscript{classes }}training and other LSC features
previously reported in~\cite{LSC}. The possibility to obtain good
performance with the reduced embedding size might also proportionally
reduce the required storage size of vector databases. This makes the
possibility to estimate \emph{n\textsubscript{min}} and use it with LSC
to train NNs a valuable property of the proposed method.

\unskip

\subsection{Desired properties of NN LS}
\label{desiredLS}

Whereas LSC allows to obtain NNs with specific LS configurations, the
complete set of desired LS properties which is beneficial for NN
training and inference is still unknown. It has previously been shown
that low inter-class and high itra-class embedding similarity is
essential for good NN performance~\cite{CenterLoss,SphereFace}. This has
led to \emph{A\textsubscript{n}} being chosen in~\cite{LSC} owing to its
sparse vector allocation. However, the \emph{CEembs} study has also
shown that perfectly uniform vector systems are harder to train due to
inherent dissimilarity between certain classes, which is reflected in
non-uniform distribution of clusters of unconstrained NNs. This
observation has led to the assumption that \emph{CEembs} is the best
possible configuration, which has been disproved in Section~\ref{i1k} of this
paper. The similarity between \emph{n\textsubscript{dim}}=384
\emph{CEembs} and \emph{n\textsubscript{min}} \(V_{n}^{21}\) suggests
that \emph{CEembs} might live on a low-dimensional manifold in
\emph{n\textsubscript{dim}} space, an observation previously reported in
literature~\cite{LLID}. In this case \emph{n\textsubscript{min}} approach
can be viewed as good estimation of the dimension of this manifold for
specific LS configurations.

The existence of universal LS, or universal embeddings, was previously
hypothesized in~\cite{Plato}. In simple terms this hypothesis suggests
that, as NNs improve, all NN embeddings eventually converge to the
embeddings in the same LS which most accurately represents the real
world. It has recently been used to successfully convert embeddings of
different NNs into one LS in unsupervised manner~\cite{HUG}.
Unfortunately, currently existing research does not provide any details
regarding what the universal LS might look like and what properties it
might have. Nevertheless, assuming such LS exists, it would be the
perfect target for NN LS configuration. Therefore, this topic is
extremely relevant to LSC research and will be studied in greater detail
in the future.

\section{Conclusions}
\label{conclusions}

This paper studies properties of vector systems which can be used as
target LS configurations for NN training. The number and spacing of
vector in the systems are used as criteria for assessing the vector
system applicability. It is shown that by adding additional elements to
\emph{A\textsubscript{n}} base vector one obtains significantly more
vectors with an acceptable decrease in their spacing. It is shown that
training with said systems allows faster convergence compared to
\emph{A\textsubscript{n}} and interpolated \emph{A\textsubscript{n}}
training. It is also shown that using minimum LS dimension which allows
to incorporate the desired number of classes significantly speeds up LSC
training. This is verified using 50k, 300k, and 600k large
\emph{n\textsubscript{classes}} dataset training along with conventional
i1k 1000 classes training. The latter experiment shows that training
with LSC is still slower than the conventional classifier training with
CE loss, but the gap between the two methods has been significantly
reduced. The results also suggest that LSC allows to estimate the
necessary dimension of manifold which accommodates NN embeddings even
when training with conventional methods. Furthermore, the possibility to
use NNs with reduced embedding dimensions has potential benefits for
embedding vector database size reduction.

\section*{Acknowledgement}
\label{acknowledgement}

The author would like to thank his colleagues Dr Igor V. Netay, Dr Anton Raskovalov, 
and Ilya Androsov for fruitful discussions, and Vasily Dolmatov for discussions and project supervision.


\bibliographystyle{IEEEtran}
\bibliography{IEEEabrv,ms}

%% file: ms.bbl
\begin{thebibliography}{10}
\providecommand{\url}[1]{#1}
\csname url@samestyle\endcsname
\providecommand{\newblock}{\relax}
\providecommand{\bibinfo}[2]{#2}
\providecommand{\BIBentrySTDinterwordspacing}{\spaceskip=0pt\relax}
\providecommand{\BIBentryALTinterwordstretchfactor}{4}
\providecommand{\BIBentryALTinterwordspacing}{\spaceskip=\fontdimen2\font plus
\BIBentryALTinterwordstretchfactor\fontdimen3\font minus
  \fontdimen4\font\relax}
\providecommand{\BIBforeignlanguage}[2]{{%
\expandafter\ifx\csname l@#1\endcsname\relax
\typeout{** WARNING: IEEEtran.bst: No hyphenation pattern has been}%
\typeout{** loaded for the language `#1'. Using the pattern for}%
\typeout{** the default language instead.}%
\else
\language=\csname l@#1\endcsname
\fi
#2}}
\providecommand{\BIBdecl}{\relax}
\BIBdecl

\bibitem{clip}
\BIBentryALTinterwordspacing
S.~Li, L.~Sun, and Q.~Li, ``Clip-reid: Exploiting vision-language model for
  image re-identification without concrete text labels,'' 2023. [Online].
  Available: \url{https://arxiv.org/abs/2211.13977}
\BIBentrySTDinterwordspacing

\bibitem{Multimodal}
\BIBentryALTinterwordspacing
C.~Akkus, L.~Chu, V.~Djakovic, S.~Jauch-Walser, P.~Koch, G.~Loss, C.~Marquardt,
  M.~Moldovan, N.~Sauter, M.~Schneider, R.~Schulte, K.~Urbanczyk,
  J.~Goschenhofer, C.~Heumann, R.~Hvingelby, D.~Schalk, and M.~Aßenmacher,
  ``Multimodal deep learning,'' 2023. [Online]. Available:
  \url{https://arxiv.org/abs/2301.04856}
\BIBentrySTDinterwordspacing

\bibitem{CenterLoss}
Y.~Wen, K.~Zhang, Z.~Li, and Y.~Qiao, ``A discriminative feature learning
  approach for deep face recognition,'' in \emph{ECCV 2016}, 2016, pp.
  499--515.

\bibitem{lsconf}
\BIBentryALTinterwordspacing
N.~Gabdullin, ``Latent space configuration for improved generalization in
  supervised autoencoder neural networks,'' 2025. [Online]. Available:
  \url{https://arxiv.org/abs/2402.08441}
\BIBentrySTDinterwordspacing

\bibitem{SphereFace}
\BIBentryALTinterwordspacing
W.~Liu, Y.~Wen, Z.~Yu, M.~Li, B.~Raj, and L.~Song, ``Sphereface: Deep
  hypersphere embedding for face recognition,'' 2018. [Online]. Available:
  \url{https://arxiv.org/abs/1704.08063}
\BIBentrySTDinterwordspacing

\bibitem{LSC}
\BIBentryALTinterwordspacing
N.~Gabdullin, ``Using predefined vector systems as latent space configuration
  for neural network supervised training on data with arbitrarily large number
  of classes,'' 2025. [Online]. Available:
  \url{https://arxiv.org/abs/2510.04090}
\BIBentrySTDinterwordspacing

\bibitem{Igor}
\BIBentryALTinterwordspacing
I.~V. Netay, ``Series of quasi-uniform scatterings with fast search, root
  systems and neural network classifications,'' 2025. [Online]. Available:
  \url{https://arxiv.org/abs/2512.04865}
\BIBentrySTDinterwordspacing

\bibitem{Lie}
J.~E. Humphreys, \emph{Introduction to Lie algebras and representation theory},
  ser. Graduate texts in mathematics.\hskip 1em plus 0.5em minus 0.4em\relax
  New York, NY: Springer-Verlag, 1972, vol.~9.

\bibitem{LieART}
\BIBentryALTinterwordspacing
R.~Feger, T.~W. Kephart, and R.~J. Saskowski, ``Lieart 2.0 – a mathematica
  application for lie algebras and representation theory,'' \emph{Computer
  Physics Communications}, vol. 257, p. 107490, Dec. 2020. [Online]. Available:
  \url{http://dx.doi.org/10.1016/j.cpc.2020.107490}
\BIBentrySTDinterwordspacing

\bibitem{LLID}
\BIBentryALTinterwordspacing
C.~Li, H.~Farkhoor, R.~Liu, and J.~Yosinski, ``Measuring the intrinsic
  dimension of objective landscapes,'' 2018. [Online]. Available:
  \url{https://arxiv.org/abs/1804.08838}
\BIBentrySTDinterwordspacing

\bibitem{Plato}
\BIBentryALTinterwordspacing
M.~Huh, B.~Cheung, T.~Wang, and P.~Isola, ``The platonic representation
  hypothesis,'' 2024. [Online]. Available:
  \url{https://arxiv.org/abs/2405.07987}
\BIBentrySTDinterwordspacing

\bibitem{HUG}
\BIBentryALTinterwordspacing
R.~Jha, C.~Zhang, V.~Shmatikov, and J.~X. Morris, ``Harnessing the universal
  geometry of embeddings,'' 2025. [Online]. Available:
  \url{https://arxiv.org/abs/2505.12540}
\BIBentrySTDinterwordspacing

\end{thebibliography}
